\documentclass[letter, 10 pt, conference]{ieeeconf}
\pdfoutput=1
\IEEEoverridecommandlockouts                              
\usepackage{graphicx, subfigure, float, placeins}
\usepackage{amssymb, amsthm, amsfonts, amsmath, mathrsfs, textcomp, mathtools, gensymb, stmaryrd}
\usepackage{array, tabularx, booktabs, multirow, makecell}
\usepackage{xfrac}

\usepackage[font=small,labelfont=bf]{caption}

\usepackage[ruled,vlined]{algorithm2e}

\SetAlFnt{\small}

\usepackage{tikz, tikz-3dplot, pgfplots}
\usetikzlibrary{snakes, decorations, shapes, arrows, chains, positioning, shapes.symbols, patterns}
\pgfdeclarelayer{background}
\pgfdeclarelayer{foreground}
\pgfsetlayers{background,main,foreground}

\newcommand{\etal}{et~al.}

\newcommand{\ie}{\emph{i.e.,}}

\newcommand*\rot{\rotatebox{90}}
\newcolumntype{L}[1]{>{\raggedright\let\newline\\\arraybackslash\hspace{0pt}}m{#1}}
\newcolumntype{M}[1]{>{\centering\let\newline\\\arraybackslash\hspace{0pt}}m{#1}}
\newcolumntype{R}[1]{>{\raggedleft\let\newline\\\arraybackslash\hspace{0pt}}m{#1}}
\newcommand{\block}[2]{\rot{\shortstack[l]{\tikz \draw[color=black, fill=#1] (0,0) rectangle ++(0.25,0.25) {}; \\ {#2}}}}

\makeatletter
\let\NAT@parse\undefined
\makeatother

\usepackage{footnote}

\usepackage[hidelinks=true, bookmarks=false, pdfauthor={summer.icequeen}]{hyperref} 
\hypersetup{colorlinks=true, linkcolor=blue, citecolor=red, filecolor=cyan, urlcolor=blue}
\usepackage[noadjust]{cite}

\usepackage[capitalize, nameinlink, noabbrev]{cleveref}

\pgfplotsset{compat=1.9}
\usepackage{balance}

\usepackage{xcolor}

\definecolor{grayd}{RGB}{60,60,60}
\definecolor{grayl}{RGB}{160,160,160}

\definecolor{viridis0}{HTML}{481567}
\definecolor{viridis1}{HTML}{453781}
\definecolor{viridis2}{HTML}{39568C}
\definecolor{viridis3}{HTML}{2D708E}
\definecolor{viridis4}{HTML}{238A8D}
\definecolor{viridis5}{HTML}{20A387}
\definecolor{viridis6}{HTML}{3DBB75}
\definecolor{viridis7}{HTML}{73D055}
\definecolor{viridis8}{HTML}{B8DE29}
\definecolor{viridis9}{HTML}{FDE728}

\definecolor{rainbow1}{HTML}{FF5454}
\definecolor{rainbow2}{HTML}{FF8000}
\definecolor{rainbow3}{HTML}{FDE728}
\definecolor{rainbow4}{HTML}{C8E64C}
\definecolor{rainbow5}{HTML}{8CD446}
\definecolor{rainbow6}{HTML}{45D2B0}
\definecolor{rainbow7}{HTML}{438CCB}
\definecolor{rainbow8}{HTML}{4262FE}
\definecolor{rainbow9}{HTML}{5240C3}
\definecolor{rainbow10}{HTML}{7C3FC0}
\definecolor{rainbow11}{HTML}{D145C1}

\definecolor{icra1}{HTML}{FDE728}   
\definecolor{icra2}{HTML}{C8E64C}   
\definecolor{icra3}{HTML}{8CD446}   
\definecolor{icra4}{HTML}{4DC742}   
\definecolor{icra5}{HTML}{45D2B0}   
\definecolor{icra6}{HTML}{438CCB}   
\definecolor{icra7}{HTML}{4262FE}   
\definecolor{icra8}{HTML}{5240C3}   
\definecolor{icra9}{HTML}{7C3FC0}   
\definecolor{icra10}{HTML}{D145C1}  
\definecolor{icra11}{HTML}{FF5454}  
\definecolor{icra12}{HTML}{FF8000}  
\definecolor{icra13}{HTML}{FFA054}  

\definecolor{cnn0}{HTML}{440154}
\definecolor{cnn1}{HTML}{404788}
\definecolor{cnn2}{HTML}{20A387}
\definecolor{cnn3}{HTML}{287D8E}
\definecolor{colorcnn}{HTML}{440154}
\definecolor{colorwarp}{HTML}{404788}

\definecolor{city0}{RGB}{128, 64,128}
\definecolor{city1}{RGB}{244, 35,232}
\definecolor{city2}{RGB}{ 70, 70, 70}
\definecolor{city3}{RGB}{102,102,156}
\definecolor{city4}{RGB}{190,153,153}
\definecolor{city5}{RGB}{153,153,153}
\definecolor{city6}{RGB}{250,170, 30}
\definecolor{city7}{RGB}{220,220,  0}
\definecolor{city8}{RGB}{107,142, 35}
\definecolor{city9}{RGB}{152,251,152}
\definecolor{city10}{RGB}{ 70,130,180}
\definecolor{city11}{RGB}{220, 20, 60}
\definecolor{city12}{RGB}{255,  0,  0}
\definecolor{city13}{RGB}{  0,  0,142}
\definecolor{city14}{RGB}{  0,  0, 70}
\definecolor{city15}{RGB}{  0, 60,100}
\definecolor{city16}{RGB}{  0, 80,100}
\definecolor{city17}{RGB}{  0,  0,230}
\definecolor{city18}{RGB}{119, 11, 32}

\definecolor{haar1}{HTML}{453781} 
\definecolor{haar2}{HTML}{2D708E} 
\definecolor{haar3}{HTML}{20A387} 
\definecolor{haar4}{HTML}{FDE728} 
\definecolor{haar5}{HTML}{6699CC} 

\newcommand{\iali}[1]{\begin{align}#1\end{align}}
\newcommand{\ialid}[1]{\begin{aligned}#1\end{aligned}}
\newcommand{\ieqn}[1]{\begin{equation}#1\end{equation}}

\newcommand{\tY}{\tilde{Y}}

\renewcommand{\tilde}{\widetilde}

\graphicspath{{./figures/}}

\title{\LARGE\bfseries Detailed Dense Inference with Convolutional Neural Networks \\via Discrete Wavelet Transform}
\author{Lingni Ma$^1$, J{\"o}rg St\"uckler$^2$, Tao Wu$^1$ and Daniel Cremers$^1$%
\thanks{$^{1}$ Lingni Ma, Tao Wu and Daniel Cremers are with the Computer Vision and Artificial Intelligence Group at the Computer Science Department, Technical University of Munich, ({\tt\small \{lingni,tao.wu,cremers\}@in.tum.de})} %
\thanks{$^2$ J{\"o}rg St\"uckler ({\tt\small joerg.stueckler@tuebingen.mpg.de}) is with Max Planck Institute for Intelligent Systems.}%
}

\begin{document}
\maketitle

\thispagestyle{empty}
\pagestyle{empty}

\begin{abstract}
Dense pixelwise prediction such as semantic segmentation is an up-to-date challenge for deep convolutional neural networks (CNNs). Many state-of-the-art approaches either tackle the loss of high-resolution information due to pooling in the encoder stage, or use dilated convolutions or high-resolution lanes to maintain detailed feature maps and predictions. Motivated by the structural analogy between multi-resolution wavelet analysis and the pooling/unpooling layers of CNNs, we introduce discrete wavelet transform (DWT) into the CNN encoder-decoder architecture and propose WCNN. The high-frequency wavelet coefficients are computed at encoder, which are later used at the decoder to unpooled jointly with coarse-resolution feature maps through the inverse DWT. The DWT/iDWT is further used to develop two wavelet pyramids to capture the global context, where the multi-resolution DWT is applied to successively reduce the spatial resolution and increase the receptive field. Experiment with the Cityscape dataset, the proposed WCNNs are computationally efficient and yield improvements the accuracy for high-resolution dense pixelwise prediction.

\end{abstract}

\section{Introduction}


Dense pixelwise prediction tasks such as semantic segmentation, optical flow or depth estimation remain up-to-date challenges in computer vision. They find rapidly rising interests for applications such as autonomous driving, robotic vision and image scene understanding. Succeeded by its remarkable success in image recognition~\cite{cnn:krizhevsky12nips:alexnet}, deep convolutional neural networks (CNNs) have achieved state-of-the-art performances in dense prediction tasks such as semantic segmentation~\cite{cnn:zhao17cvpr:pspnet,cnn:lin17cvpr:refinenet,cnn:pohlen17cvpr:frrn} or single-image depth estimation~\cite{cnn:laina163dv:depth}.

Many dense prediction tasks consist of two concurrent goals: classification and localization. Classification is well tackled by an end-to-end trainable CNN architecture, e.g.~VGGNet~\cite{cnn:simonyan15iclr:vgg} or ResNet~\cite{cnn:he16cvpr:resnet}, which typically stacks multiple layers of successive convolution, nonlinear activation, and pooling. A typical pooling step, which performs either a subsampling or some strided averaging on an input volume, is favorable for the invariance of prediction results to small spatial translations in the input data as well as for the boost of computational efficiency via dimension reduction. Its downside, however, is the loss of resolution in output feature maps, which renders high-quality pixelwise prediction challenging.

Several remedies for such a dilemma have been proposed in the literatures. As suggested in \cite{cnn:noh15iccv:deconv,cnn:badrinarayanan15:segnet}, one may mirror the encoder network by a decoder network. Each upsampling (or unpooling) layer in the decoder network is introduced in symmetry to a corresponding pooling layer in the encoder network, and then followed by trainable convolutional layers. Alternatively, one may use dilated (also known as atrous) convolutions in a CNN encoder as proposed in \cite{cnn:yu16iclr:dilate, cnn:chen15iclr:deeplab, cnn:chen18:deeplabv2}. This enables the CNN to expand the receptive fields of pixels as convolutional layers stack up without losing resolution in the feature maps, however, at the cost of significant computational time and memory. Another alternative is to combine a CNN low-resolution classifier with a conditional random field (CRF) \cite{misc:krahenb11nips:crf, cnn:krahenbuhl13icml:param}, either as a stand-alone post-processing step \cite{cnn:chen15iclr:deeplab, cnn:chen18:deeplabv2} or combined with a CNN in an end-to-end trainable architecture \cite{cnn:zheng15iccv:crfrnn, cnn:lin16cvpr:piecewise}. The latter also comes with an increased demand in run-time for training and inference.

Motivated by close analogy between pooling (resp.~unpooling) in an encoder-decoder CNN and decomposition (resp.~reconstruction) in multi-resolution wavelet analysis, this paper proposes a new class of CNNs with wavelet unpooling and wavelet pyramid. We name the network WCNN. The first contribution with WCNN is to achieve unpooling with the inverse discrete wavelet transform (iDWT). To this end, DWT is applied at the encoder to decompose feature maps into frequency bands. The high frequency components are skip-connected to the decoder to perform iDWT jointly with the coarse-resolution feature maps. The wavelet unpooling does not require any additional parameters over baseline CNNs, where the overhead only comes from the memory to cache frequency  coefficients from encoder. The second contribution of WCNN are two wavelet-based pyramid variants to bridge the standard encoder and decoder. The wavelet pyramids obtain global context from a receptive field of the entire image by exploiting multi-resolution DWT/iDWT. The experiments over the dataset Cityscape show that the proposed WCNN yields systematically improvements in dense prediction accuracy.


\section{Related Work}
Many challenging tasks in computer vision such as single image depth prediction or semantic image segmentation require models for dense prediction, since they either involve regressing quantities pixelwise or classifying the pixels. Most current state-of-the-art methods for dense prediction tasks are based on end-to-end trainable deep learning architectures. Early methods segment the image into regions such as superpixels in a bottom-up fashion. Predictions for the regions are determined  based on deep neural network features \cite{misc:yan15cvpr:object,misc:farabet13pami:hierfeatscenelabeling,cnn:liu15cvpr:deepneuralfields}. The use of image-based bottom-up regions supports adherence of the dense predictions to the boundaries in the image.

Aim at end-to-end CNNs, Long~\etal~\cite{cnn:long15cvpr:fcn} propose a fully connected convolutional (FCN) architecture for semantic image segmentation which successively convolves and pools feature maps of an increasing number of feature channels. FCNs employ the transposed convolution to learn the upsampling of coarse feature maps. To obtain segmentation, feature maps of the intermediate resolutions are concatenated and further processed by transposed convolutions. Since the introduction of FCNs, many variants for dense prediction are proposed. Hariharan~\etal~\cite{cnn:hariharan15cvpr:hypercolumns} classify pixels based on feature vectors that are extracted at corresponding locations across all feature maps in a CNN. This way, the method combines features across all layers available in the network, capturing high-resolution detail as well as context in large receptive fields. However, this approach becomes inefficient in deep architectures with many wide layers. Noh~\etal~\cite{cnn:noh15iccv:deconv} and Dosovitsky~\etal~\cite{cnn:dosovitskiy15iccv:flownet} propose encoder-decoder CNN architectures which successively unpool and convolve the lowest resolution feature map of the encoder back to a high output resolution. Since the feature maps in the encoder lose spatial information through pooling, Noh~\etal~\cite{cnn:noh15iccv:deconv} exploint the memorized unpooling~\cite{cnn:zeiler11cvpr:memorizedunpool} to upscale coarse feature maps at the decoder stage, where the pooling locations are used to unpool accordingly. The FCN of Laina~\etal~\cite{cnn:laina163dv:depth} uses the deep residual network~\cite{cnn:he16cvpr:resnet} as an encoder, where most pooling layers are replaced by stride-two convolution. For upscaling, the upprojection block is developed as an efficient implementation of upconvolution. The principle of upconvolution is developed by \cite{cnn:dosovitskiy15cvpr:chairs}, which first unpools a feature map by putting activations to one entry of a $2\times 2$ block and then filter the sparse feature map with convolution. Details in the predictions of such encoder-decoder FCNs can be improved by feeding the feature maps in each scale of the encoder to the corresponding scale of the decoder (skip connections, e.g.~\cite{cnn:dosovitskiy15iccv:flownet}). In RefineNet~\cite{cnn:lin17cvpr:refinenet}, the decoder feature maps are successively refined using multi-resolution fusion with their higher resolution counterparts in the encoder. In this paper, we also reincorporate the high-frequency information that is discarded during pooling to successively refine feature maps in the decoder.

Some FCN architectures use dilated convolutions in order to increase receptive field without pooling and maintain high-resolution of the feature maps~\cite{cnn:chen15iclr:deeplab, cnn:chen18:deeplabv2, cnn:yu16iclr:dilate}. These dilated CNNs trade high-resolution output with the high memory consumption, which quickly become a bottleneck for training with large batch size for encoder-decoder CNNs. The full-resolution residual network (FRRN) by \cite{cnn:pohlen17cvpr:frrn} is an alternative model which keeps features in a high-resolution lane and at the same time, extracts low-resolution higher-order features in an encoder-decoder architecture. The high-resolution features are successively refined from residuals computed through the encoder-decoder lane. While the model is highly demanding in memory and training time, it achieves high-resolution predictions that well adhere to segment boundaries.
\cite{cnn:ghiasi16eccv:lrr} take inspiration from Laplace image decompositions for their network design. They successively refine the high-frequency parts of the score maps in order to improve predictions at segment boundaries. Structured prediction approaches integrate inference in CRFs with deep neural networks in end-to-end trainable models~\cite{cnn:zheng15iccv:crfrnn, cnn:liu15iccv:semantic, cnn:chandra16eccv:gaussiancrfs, cnn:lin16cvpr:piecewise}. While the models are capable of recovering high-resolution predictions, inference and learning typically requires tedious iterative procedures. In contrast to those approaches, we aim to provide detailed predictions in a swift and direct forward pass. Recently, the pyramid scene parsing network (PSPNet) from \cite{cnn:zhao17cvpr:pspnet} extracts global context features using a pyramid pooling module, which shows the benefit of aggregation global information for dense predictions. The pyramid design in PSPNet relies multiple average pooling layers with heuristic window size. In this work, we also propose a more efficient pyramid pooling stage based on multi-resolution DWT.

\section{WCNN Encoder-Decoder Architectures}
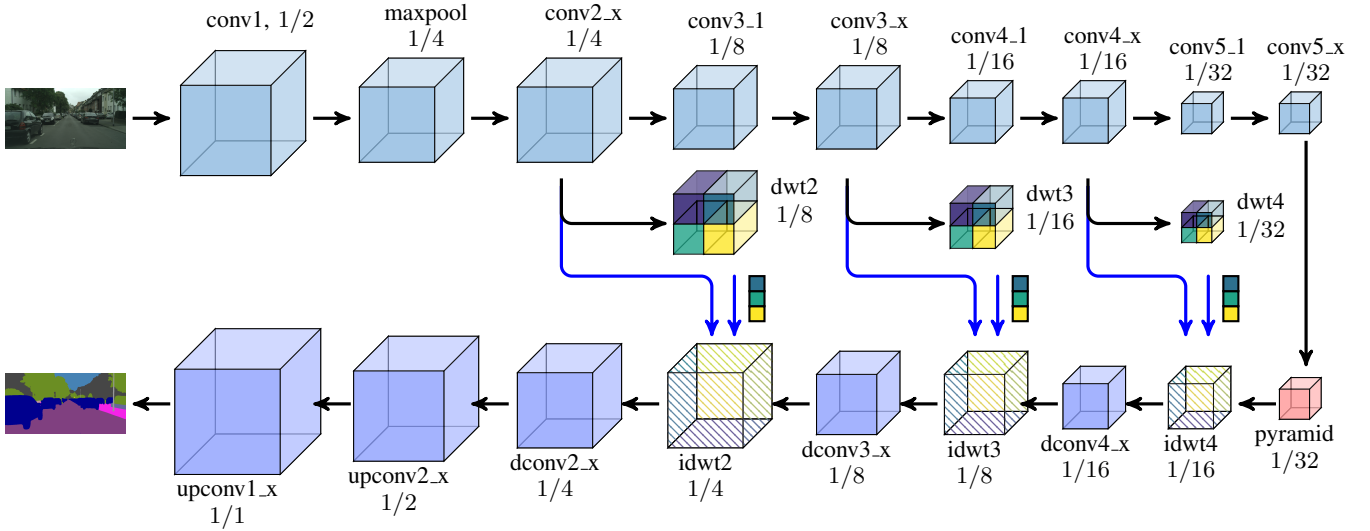
\begin{figure*}
  \begin{tikzpicture}[inner sep=0pt]

  \tikzstyle{cmt}=[draw=none, text=black, font=\small, text centered, text width =2cm, minimum height=4ex]
  \tikzstyle{arr}=[color=black, ->,>=stealth', line width=1.2pt, line cap=rounded, rounded corners]
  \tikzstyle{revarr}=[color=black, <-,>=stealth', line width=1.2pt, line cap=rounded, rounded corners]

  \def\hbxx{0.2}
  \def\lowf#1#2{
  \begin{scope}[xshift=#1, yshift=#2, line cap=round, thick]
  \draw[draw=black, fill=haar1] (0, \hbxx) -- ++(\hbxx,0) -- ++(0,\hbxx) -- ++(-\hbxx,0) -- cycle;
  \end{scope}
  }

  \def\highf#1#2{
  \begin{scope}[xshift=#1, yshift=#2, line cap=round, thick]
    \draw[arr, color=blue](-0.2, \hbxx*4) --++(0, -\hbxx*4);
    \draw[draw=black, fill=haar4] (0, \hbxx) -- ++(\hbxx,0) -- ++(0,\hbxx) -- ++(-\hbxx,0) -- cycle;
    \draw[draw=black, fill=haar3] (0, \hbxx*2) -- ++(\hbxx,0) -- ++(0,\hbxx) -- ++(-\hbxx,0) -- cycle;
    \draw[draw=black, fill=haar2] (0, \hbxx*3) -- ++(\hbxx,0) -- ++(0,\hbxx) -- ++(-\hbxx,0) -- cycle;

  \end{scope}
  }

  \newcommand\layercube[6]{
    \def\xval{#1}
    \def\yval{#2}
    \def\zval{#3}
    \coordinate (O) at (0,0,0);
    \coordinate (A) at (0,\yval,0);
    \coordinate (B) at (0,\yval,\zval);
    \coordinate (C) at (0,0,\zval);
    \coordinate (D) at (\xval,0,0);
    \coordinate (E) at (\xval,\yval,0);
    \coordinate (F) at (\xval,\yval,\zval);
    \coordinate (G) at (\xval,0,\zval);

    \draw[black,fill=none] (O) -- (C) -- (G) -- (D) -- cycle;
    \draw[black,fill=none] (O) -- (A) -- (E) -- (D) -- cycle;
    \draw[black,fill=none] (O) -- (A) -- (B) -- (C) -- cycle;
    \draw[black,fill=#4!40,opacity=0.8] (D) -- (E) -- (F) -- (G) -- cycle;
    \draw[black,fill=#4!40,opacity=0.6] (A) -- (B) -- (F) -- (E) -- cycle;
    \draw[black,fill=#4!60,opacity=0.8] (C) -- (B) -- (F) -- (G) -- cycle;
    \node at(A) [cmt, shift={(\xval*0.5, 0.4)}]{#5};
    \node at(C) [cmt, shift={(\xval*0.5,-0.4)}]{#6};

  }

  \newcommand\dwtcube[5]{
    \def\xval{#1*0.5}
    \def\yval{#2*0.5}
    \def\zval{#3}
    \node at(#1, #2*0.5, 0)   [cmt, shift={(0.5, 0.2)}] {#4};
    \node at(#1, #2*0.5, 0)   [cmt, shift={(0.5,-0.2)}] {#5};

    \begin{scope}[yshift=#2*0.5cm]
    \coordinate (O) at (0,0,0);
    \coordinate (A) at (0,\yval,0);
    \coordinate (B) at (0,\yval,\zval);
    \coordinate (C) at (0,0,\zval);
    \coordinate (D) at (\xval,0,0);
    \coordinate (E) at (\xval,\yval,0);
    \coordinate (F) at (\xval,\yval,\zval);
    \coordinate (G) at (\xval,0,\zval);
    \draw[black,fill=haar1!80,opacity=0.3] (O) -- (C) -- (G) -- (D) -- cycle;
    \draw[black,fill=haar1!80,opacity=0.3] (O) -- (A) -- (E) -- (D) -- cycle;
    \draw[black,fill=haar1!80,opacity=0.1] (O) -- (A) -- (B) -- (C) -- cycle;
    \draw[black,fill=haar1!80,opacity=0.5] (D) -- (E) -- (F) -- (G) -- cycle;
    \draw[black,fill=haar1!80,opacity=0.8] (A) -- (B) -- (F) -- (E) -- cycle;
    \draw[black,fill=haar1,opacity=0.95] (C) -- (B) -- (F) -- (G) -- cycle;
    \end{scope}

    \begin{scope}[yshift=#2*0.5cm, xshift=#1*0.5cm]
      \coordinate (O) at (0,0,0);
      \coordinate (A) at (0,\yval,0);
      \coordinate (B) at (0,\yval,\zval);
      \coordinate (C) at (0,0,\zval);
      \coordinate (D) at (\xval,0,0);
      \coordinate (E) at (\xval,\yval,0);
      \coordinate (F) at (\xval,\yval,\zval);
      \coordinate (G) at (\xval,0,\zval);
      \draw[black,fill=haar2!10,opacity=0.2] (O) -- (C) -- (G) -- (D) -- cycle;
      \draw[black,fill=haar2!20,] (O) -- (A) -- (E) -- (D) -- cycle;
      \draw[black,fill=haar2!10,opacity=0.2] (O) -- (A) -- (B) -- (C) -- cycle;
      \draw[black,fill=haar2!40,opacity=0.6] (D) -- (E) -- (F) -- (G) -- cycle;
      \draw[black,fill=haar2!40,opacity=0.6] (A) -- (B) -- (F) -- (E) -- cycle;
      \draw[black,fill=haar2,opacity=0.95] (C) -- (B) -- (F) -- (G) -- cycle;
    \end{scope}

    \coordinate (O) at (0,0,0);
    \coordinate (A) at (0,\yval,0);
    \coordinate (B) at (0,\yval,\zval);
    \coordinate (C) at (0,0,\zval);
    \coordinate (D) at (\xval,0,0);
    \coordinate (E) at (\xval,\yval,0);
    \coordinate (F) at (\xval,\yval,\zval);
    \coordinate (G) at (\xval,0,\zval);
    \draw[black,fill=none,opacity=0.2] (O) -- (C) -- (G) -- (D) -- cycle;
    \draw[black,fill=none] (O) -- (A) -- (E) -- (D) -- cycle;
    \draw[black,fill=none] (O) -- (A) -- (B) -- (C) -- cycle;
    \draw[black,fill=none,opacity=0.8] (D) -- (E) -- (F) -- (G) -- cycle;
    \draw[black,fill=none,opacity=0.2] (A) -- (B) -- (F) -- (E) -- cycle;
    \draw[black,fill=haar3,opacity=0.8] (C) -- (B) -- (F) -- (G) -- cycle;

    \begin{scope}[xshift=#1*0.5cm]
      \coordinate (O) at (0,0,0);
      \coordinate (A) at (0,\yval,0);
      \coordinate (B) at (0,\yval,\zval);
      \coordinate (C) at (0,0,\zval);
      \coordinate (D) at (\xval,0,0);
      \coordinate (E) at (\xval,\yval,0);
      \coordinate (F) at (\xval,\yval,\zval);
      \coordinate (G) at (\xval,0,\zval);
      \draw[black,fill=none] (O) -- (C) -- (G) -- (D) -- cycle;
      \draw[black,fill=none] (O) -- (A) -- (E) -- (D) -- cycle;
      \draw[black,fill=haar4!10, opacity=0.1] (O) -- (A) -- (B) -- (C) -- cycle;
      \draw[black,fill=haar4!40,opacity=0.8] (D) -- (E) -- (F) -- (G) -- cycle;
      \draw[black,fill=haar4,opacity=0.1] (A) -- (B) -- (F) -- (E) -- cycle;
      \draw[black,fill=haar4,opacity=0.8] (C) -- (B) -- (F) -- (G) -- cycle;
    \end{scope}
  }

  \newcommand\idwtcube[5]{
    \def\xval{#1}
    \def\yval{#2}
    \def\zval{#3}
    \coordinate (O) at (0,0,0);
    \coordinate (A) at (0,\yval,0);
    \coordinate (B) at (0,\yval,\zval);
    \coordinate (C) at (0,0,\zval);
    \coordinate (D) at (\xval,0,0);
    \coordinate (E) at (\xval,\yval,0);
    \coordinate (F) at (\xval,\yval,\zval);
    \coordinate (G) at (\xval,0,\zval);
    \node at(A) [cmt, shift={(\xval*0.5, 0.4)}]{#4};
    \node at(C) [cmt, shift={(\xval*0.5,-0.4)}]{#5};

    \draw[black,pattern=north west lines, pattern color=haar1] (O) -- (C) -- (G) -- (D) -- cycle;
    \draw[black,pattern=north west lines, pattern color=haar4] (O) -- (A) -- (E) -- (D) -- cycle;
    \draw[black,pattern=north west lines, pattern color=haar2] (O) -- (A) -- (B) -- (C) -- cycle;
    \draw[black,pattern=north west lines, pattern color=haar2!60,opacity=0.8] (D) -- (E) -- (F) -- (G) -- cycle;
    \draw[black,pattern=north west lines, pattern color=haar3!60,opacity=0.6] (A) -- (B) -- (F) -- (E) -- cycle;
    \draw[black,pattern=north west lines, pattern color=haar1!40,opacity=0.6] (C) -- (B) -- (F) -- (G) -- cycle;
  }

  \node(rgb) at (-2.8cm,-0.3cm)[anchor=west]{\includegraphics[width=1.6cm]{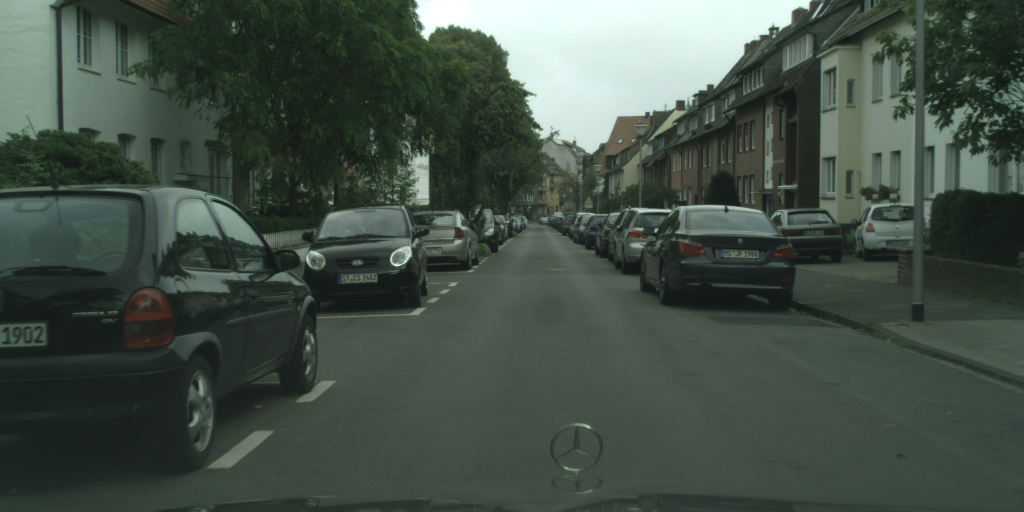}};

  \begin{scope}[yshift=-0.6cm] 
    \layercube{1.2}{1.2}{1.2}{icra6}{conv1, $1/2$}{}
  \end{scope}

  \begin{scope}[xshift=2.3cm,yshift=-0.5cm] 
    \layercube{1.0}{1.0}{1.0}{icra6}{maxpool\\$1/4$}{}
  \end{scope}

  \begin{scope}[xshift=4.4cm,yshift=-0.5cm] 
    \layercube{1.0}{1.0}{1.0}{icra6}{conv2\_x\\$1/4$}{}
  \end{scope}

  \begin{scope}[xshift=6.4cm,yshift=-0.4cm] 
    \layercube{0.8}{0.8}{0.8}{icra6}{conv3\_1\\$1/8$}{}
  \end{scope}
  \begin{scope}[xshift=6.4cm,yshift=-1.8cm] 
    \dwtcube{0.8}{0.8}{0.8}{dwt2}{$1/8$};
  \end{scope}
  \begin{scope}[xshift=8.3cm,yshift=-0.4cm] 
    \layercube{0.8}{0.8}{0.8}{icra6}{conv3\_x\\$1/8$}{}
  \end{scope}

  \begin{scope}[xshift=10.0cm,yshift=-0.4cm] 
    \layercube{0.6}{0.6}{0.6}{icra6}{conv4\_1\\$1/16$}{}
  \end{scope}
  \begin{scope}[xshift=10.0cm,yshift=-1.8cm] 
    \dwtcube{0.6}{0.6}{0.6}{dwt3}{$1/16$};
  \end{scope}

  \begin{scope}[xshift=11.5cm,yshift=-0.4cm] 
    \layercube{0.6}{0.6}{0.6}{icra6}{conv4\_x\\$1/16$}{}
  \end{scope}
  \begin{scope}[xshift=13.0cm,yshift=-0.35cm] 
    \layercube{0.4}{0.4}{0.4}{icra6}{conv5\_1\\ $1/32$}{}
  \end{scope}
  \begin{scope}[xshift=13.0cm,yshift=-1.8cm] 
    \dwtcube{0.4}{0.4}{0.4}{dwt4}{$1/32$};
  \end{scope}

  \begin{scope}[xshift=14.3cm,yshift=-0.35cm] 
    \layercube{0.4}{0.4}{0.4}{icra6}{conv5\_x\\$1/32$}{}
  \end{scope}
  %
  %
  \highf{7.10cm} {-3.2cm}
  \highf{10.6cm}{-3.2cm}
  \highf{13.4cm}{-3.2cm}

  \draw[arr, draw=blue] (4.6,-1.2)--++(0,-1.2) --++(2.0, 0) --++(0,-0.8);
  \draw[arr, draw=blue] (8.4,-1.2)--++(0,-1.2) --++(1.7, 0) --++(0,-0.8);
  \draw[arr, draw=blue] (11.6,-1.2)--++(0,-1.2) --++(1.3, 0) --++(0,-0.8);

  \draw[arr, draw=black] (4.6,-1.1)--++(0,-0.6) --++(1.4, 0);
  \draw[arr, draw=black] (8.4,-1.1)--++(0,-0.6) --++(1.3, 0);
  \draw[arr, draw=black] (11.6,-1.1)--++(0,-0.6) --++(1.1, 0);
  \draw[arr, draw=black] (-1.1,-0.3)--++(0.5,0);
  \draw[arr, draw=black] (1.3, -0.3)--++(0.5,0); 
  \draw[arr, draw=black] (3.4, -0.3)--++(0.5,0); 
  \draw[arr, draw=black] (5.5, -0.3)--++(0.5,0); 
  \draw[arr, draw=black] (7.4, -0.3)--++(0.5,0); 
  \draw[arr, draw=black] (9.2, -0.3)--++(0.5,0); 
  \draw[arr, draw=black] (10.7, -0.3)--++(0.5,0);
  \draw[arr, draw=black] (12.2, -0.3)--++(0.5,0);
  \draw[arr, draw=black] (13.5, -0.3)--++(0.5,0);
  \draw[arr, draw=black] (14.5, -0.6)--++(0.,-3.0);

  \begin{scope}[yshift=-3.8cm]
    \node(pre) at(-2.8cm,-0.3cm)[anchor=west]{\includegraphics[width=1.6cm]{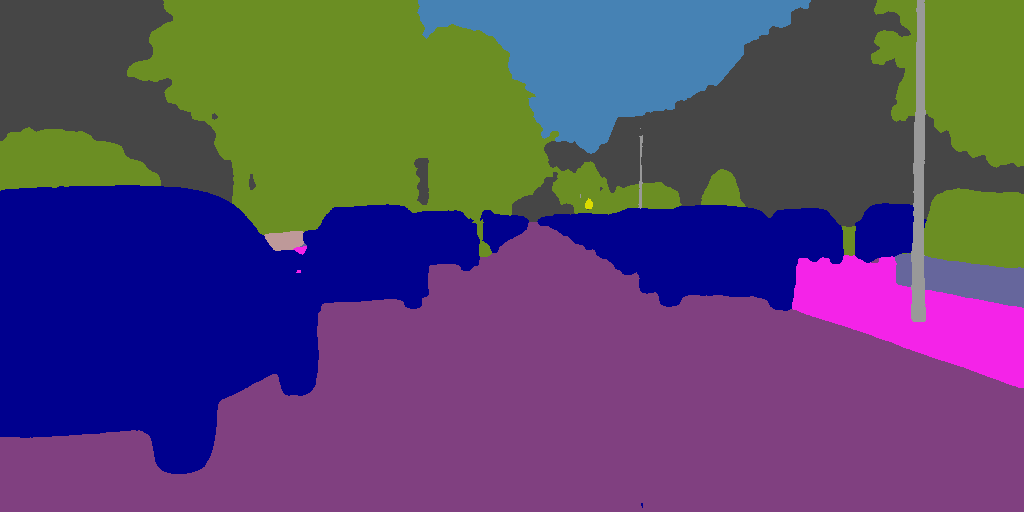}};

    \begin{scope}[yshift=-0.7cm] 
      \layercube{1.4}{1.4}{1.4}{icra7}{}{upconv1\_x\\$1/1$}
    \end{scope}

    \begin{scope}[xshift=2.3cm,yshift=-0.6cm] 
      \layercube{1.2}{1.2}{1.2}{icra7}{}{upconv2\_x\\$1/2$}
    \end{scope}

    \begin{scope}[xshift=4.4cm,yshift=-0.5cm] 
      \layercube{1.}{1.}{1.}{icra7}{}{dconv2\_x\\$1/4$}
    \end{scope}
    \begin{scope}[xshift=6.4cm,yshift=-0.5cm] 
      \idwtcube{1.}{1.}{1.}{}{idwt2\\$1/4$}
    \end{scope}

    \begin{scope}[xshift=8.3cm,yshift=-0.4cm] 
      \layercube{0.8}{0.8}{0.8}{icra7}{}{dconv3\_x\\$1/8$}
    \end{scope}
    \begin{scope}[xshift=10.0cm,yshift=-0.4cm] 
      \idwtcube{0.8}{0.8}{0.8}{}{idwt3\\$1/8$}
    \end{scope}

    \begin{scope}[xshift=11.5cm,yshift=-0.4cm] 
      \layercube{0.6}{0.6}{0.6}{icra7}{}{dconv4\_x\\$1/16$}
    \end{scope}
    \begin{scope}[xshift=12.9cm,yshift=-0.4cm] 
      \idwtcube{0.6}{0.6}{0.6}{}{idwt4\\$1/16$}
    \end{scope}

    \begin{scope}[xshift=14.3cm,yshift=-0.35cm] 
      \layercube{0.4}{0.4}{0.4}{rainbow1}{}{pyramid\\$1/32$}
    \end{scope}

    \draw[revarr, draw=black] (-1.1,-0.3)--++(0.5,0);
    \draw[revarr, draw=black] (1.3, -0.3)--++(0.5,0); 
    \draw[revarr, draw=black] (3.4, -0.3)--++(0.5,0); 
    \draw[revarr, draw=black] (5.4, -0.3)--++(0.5,0); 
    \draw[revarr, draw=black] (7.4, -0.3)--++(0.5,0); 
    \draw[revarr, draw=black] (9.1, -0.3)--++(0.5,0); 
    \draw[revarr, draw=black] (10.7, -0.3)--++(0.5,0);
    \draw[revarr, draw=black] (12.1, -0.3)--++(0.5,0);
    \draw[revarr, draw=black] (13.6, -0.3)--++(0.5,0);

  \end{scope}

  \end{tikzpicture}
  \caption{The encoder-decoder architecture of the proposed WCNN, where the data flow is indicated by black arrows and shortcuts are by blue arrows. Assume the input resolution is 1, the output resolution of each building block is denoted by $1/x$. WCNN employs ResNet \cite{cnn:he16cvpr:resnet} for the encoder, which reduces the input resolution by a factor of 32 via stride-two convolutional layers, except for one maxpool layer after conv1. To restore the input resolution, WCNN inserts three DWT layer after conv2, conv3 and conv4. The high frequencies from DWT layers are used in the decoder to perform unpooling by the iDWT layers. To extract global context, WCNN introduces two pyramid variants to bridge the encoder and decoder, which also exploits DWT/iDWT layers (see details in \cref{arxiv18:fig:pyramids}). }

  \label{arxiv18:fig:network}
\end{figure*}

Recently, CNNs have demonstrated impressive performance on many dense pixelwise prediction tasks, including image semantic segmentation, optical flow estimation, and depth regression. CNNs extract image features through successive layers of convolution and non-linear activation. In encoder architectures, as the stack of layers gets deeper, the dimension of the feature vectors increases while the spatial resolution is reduced. For dense prediction tasks, CNNs with encoder-decoder architecture are widely applied in which the feature maps of the encoder are successively unpooled and deconvolved. Research on architectures for the encoder part is relatively mature, e.g., the state-of-the-art CNNs such as~VGGNet~\cite{cnn:simonyan15iclr:vgg} and ResNet~\cite{cnn:he16cvpr:resnet} are commonly used in various applications. In contrast, the design of the decoder has not yet converged to a universally accepted solution. While it is easy to reduce spatial dimension by either pooling or strided convolution, recovering a detailed prediction from a coarse and high-dimensional feature space is less straight-forward. In this paper, we make an analogy between CNN encoder-decoders to the multi-resolution wavelet transform (see \cref{arxiv18:fig:network}). We match the pooling operations of the CNN encoder with the multilevel forward transformation of a signal by a wavelet. The decoder performs the corresponding inverse wavelet transform for unpooling. The analogy is straight-forward: the wavelet transform successively filters the signal into frequency subbands while reducing the spatial resolution. The inverse wavelet transform successively composes the frequency subband back to full resolution. While the encoder and the decoder transform between different domains (e.g. image-to-semantic segmentation vs. image-to-image in wavelet transforms), we find that wavelet unpooling provides an elegant mechanism to transmit high-frequency information from the image domain to the semantic segmentation. It also imposes a strong architectural regularization, as the feature dimensions between the encoder and the decoder need to match through the wavelet coefficients.

\subsection{Discrete Wavelet Transform}
We briefly introduce main concepts of DWT (see~\cite{book:mallat09:wavelet} for a comprehensive introduction). The multi-resolution wavelet transform provides localized time-frequency analysis of signals and images. Consider a 2D input data $X\in\mathbb{R}^{2M\times 2N}$, $\phi\in\mathbb{R}^2$ and $\psi\in\mathbb{R}^2$ as 1D low-pass and high-pass filters, respectively. Denote the indexed array element by $x_{ij}$, the single-level DWT is defined as follows,
\ieqn{\label{eq:dwt}
\ialid{
&y^{ll}_{kl} =\sum_{l}\sum_{k} x_{2i+k,2j+l}\phi_k\phi_l, \\
&y^{lh}_{kl} =\sum_{l}\sum_{k} x_{2i+k,2j+l}\phi_k\psi_l, \\
&y^{hl}_{kl} =\sum_{l}\sum_{k} x_{2i+k,2j+l}\psi_k\phi_l, \\
&y^{hh}_{kl} =\sum_{l}\sum_{k} x_{2i+k,2j+l}\psi_k\psi_l.
}}
All the convolutions above are performed with stride 2, yielding a subsampling of factor 2 along each spatial dimension. Let the low-low frequency component $Y^{ll}:=\{y^{ll}_{kl}\}$, the low-high frequency component $Y^{lh}:=\{y^{lh}_{kl}\}$, the high-low frequency component $Y^{hl}:=\{y^{hl}_{i,j}\}$, and the high-high frequency component $Y^{hh}:=\{y^{hh}_{i,j}\}$. The DWT results in $\{Y^{ll}, Y^{lh}, Y^{hl}, Y^{hh}\}\in\mathbb{R}^{M\times N}$.
Conversely, supplied with the wavelet coefficients, and provided that~$\{\phi,\psi\}$ and~$\{\tilde\phi,\tilde\psi\}$ are bi-orthogonal wavelet filters, the original input~$X$ can be reconstructed by the inverse DWT as
\iali{
x_{ij} =&\sum_{l}\sum_{k} \bigg(
y^{ll}_{kl}\tilde\phi_{i-2k}\tilde\phi_{j-2l}+
y^{lh}_{kl}\tilde\phi_{i-2k}\tilde\psi_{j-2l} \notag\\
&
+y^{hl}_{kl}\tilde\psi_{i-2k}\tilde\phi_{j-2l}+
y^{hh}_{kl}\tilde\psi_{i-2k}\tilde\psi_{j-2l}\bigg)\, .
\label{eq:idwt}
}

A cascaded wavelet decomposition successively performs \cref{eq:dwt} on low-low frequency coefficients~$\{(\cdot)^{ll}\}$ from fine to coarse resolution, while the reconstruction works reversely from coarse to fine resolution. In this sense, decomposition-reconstruction in multi-resolution wavelet analysis is in analogy to the pooling-unpooling steps in an encoder-decoder CNN (e.g., \cite{cnn:noh15iccv:deconv}). Moreover, it is worth noting that, while the low-frequency coefficients~$\{(\cdot)^{ll}\}$ store local averages of the input data, its high-frequency counterparts, namely $\{(\cdot)^{lh}\}$, $\{(\cdot)^{hl}\}$, and $\{(\cdot)^{hh}\}$
encode local textures which are vital in recovering sharp boundaries. This motivates us to make use of the high-frequency wavelet coefficients to improve the quality of unpooling during the decoder stage and, hence, improve the accuracy of CNN in pixelwise prediction.

Throughout this paper, we extensively use the Haar wavelet for its simplicity and effectiveness to boost the performances of the underlying CNN. In this scenario, the Haar filters used for decomposition, see \cref{eq:dwt}, are given by
\begin{equation}\label{nips17:eq:haarfilt}
  \phi =\left(\frac12\;,\;  \frac12\right)\;\;,\quad
  \psi =\left(\frac12\;,\; -\frac12\right)\;.
\end{equation}
The corresponding reconstruction filters in \cref{eq:idwt} are given by $\tilde\phi=2\phi$, $\tilde\psi=2\psi$, and hence the inverse transform reduces to a sum of Kronecker products (denoted with $\otimes$)
\iali{
X=&Y^{ll}\otimes \tilde\phi\,^\top \otimes\tilde\phi
+Y^{lh}\otimes \tilde\phi\,^\top \otimes\tilde\psi \notag\\
+&Y^{hl}\otimes \tilde\psi\,^\top \otimes\tilde\phi
+Y^{hh}\otimes \tilde\psi\,^\top \otimes\tilde\psi \;.
}
With CNNs, data at every layer are structured into 4D tensors, i.e., along the dimensions of the batch size, the channel number, the width and the height. To perform the wavelet transform for CNNs, we apply DWT/iDWT channelwise. Without confusion, the remaining text adopts the shorthand notations $G_h(X)$ for the Haar DWT and $G_h^{-1}(Y^{ll}, Y^{lh},Y^{hl},Y^{hh})$ for the corresponding iDWT.

\subsection{Wavelet CNN Encoder-Decoder Architecture}
We propose a CNN encoder-decoder that resembles multi-resolution wavelet decomposition and reconstruction by its pooling and unpooling operations. In addition, we introduce two pyramid variants to capture global contextual features based on the wavelet transformation.

\Cref{arxiv18:fig:network} gives an overview of the proposed WCNN architecture. WCNN employs ResNet~\cite{cnn:he16cvpr:resnet} for the encoder. In ResNet, the input resolution is successively reduced by a factor of 32 via one max-pooling layer and four stride-two convolutional layers, \ie conv1, conv3\_1, conv4\_1 and conv5\_1. In order to restore the input resolution with the decoder, WCNN inserts three DWT layer after conv2, conv3 and conv4 to decompose the corresponding feature maps into four frequency bands. The high frequencies $Y^{lh}, Y^{hl}, Y^{hh}$ are skip-connected to the decoder to perform unpooling via the iDWT layers, which we will discuss in details with \cref{sect:waveletunpool}. We add three convolutional residual block~\cite{cnn:he16cvpr:resnet} to filter the unpooled feature maps further before the next unpooling stage. As illustrated in \cref{arxiv18:fig:network}, the three iDWT layers upsample the output to $1/4$ input resolution. The full-resolution output is obtained with two upconvolutional blocks by transposed convolution. To bridge the encoder and decoder, the contextual pyramid with wavelet transformation is added. \Cref{sect:pyramid} will detail the pyramid design.

\begin{figure}
  \centering

  \tikzstyle{cmt}=[draw=none, text=black, font=\small, text centered, text width=10mm,minimum height=2ex]
  \tikzstyle{arr}=[color=black, ->,>=stealth', thick, line cap=rounded, rounded corners]

  \def\hbxx{0.2}

  \def\lowf#1#2#3{
    \begin{scope}[xshift=#1, yshift=#2, line cap=round, thick]
    \draw[draw=black, fill=haar1] (0, #3) -- ++(#3,0) -- ++(0,#3) -- ++(-#3,0) -- cycle;
    \end{scope}}

  \def\highf#1#2{
    \begin{scope}[shift={(#1)}, line cap=round, thick]
    \draw[draw=black, fill=haar2] (0, #2) -- ++(#2,0) -- ++(0,#2) -- ++(-#2,0) -- cycle;
    \draw[draw=black, fill=haar3] (#2, #2) -- ++(#2,0) -- ++(0,#2) -- ++(-#2,0) -- cycle;
    \draw[draw=black, fill=haar4] (#2*2, #2) -- ++(#2,0) -- ++(0,#2) -- ++(-#2,0) -- cycle;
    \end{scope}}

  \def\dwtlayer#1#2#3{
    \begin{scope}[shift={(#1)}, line cap=round, thick]
    \draw[draw=black, fill=haar1] (0, #2) -- ++(#2,0) -- ++(0,#2) -- ++(-#2,0) -- cycle;
    \draw[draw=black, fill=haar2] (#2, #2) -- ++(#2,0) -- ++(0,#2) -- ++(-#2,0) -- cycle;
    \draw[draw=black, fill=haar3] (0, 0) -- ++(#2,0) -- ++(0,#2) -- ++(-#2,0) -- cycle;
    \draw[draw=black, fill=haar4] (#2, 0) -- ++(#2,0) -- ++(0,#2) -- ++(-#2,0) -- cycle;
    \node at(#2,0) [cmt,yshift=-0.3, anchor=north]{#3};
    \end{scope}}

  \def\idwtlayer#1#2#3{
    \begin{scope}[shift={(#1)}, line cap=round, thick]
    \fill[pattern=north west lines, pattern color=haar1] (0, #2) -- ++(#2,0) -- ++(0,#2) -- ++(-#2,0) -- cycle;
    \fill[pattern=north west lines, pattern color=haar2] (#2, #2) -- ++(#2,0) -- ++(0,#2) -- ++(-#2,0) -- cycle;
    \fill[pattern=north west lines, pattern color=haar3] (0, 0) -- ++(#2,0) -- ++(0,#2) -- ++(-#2,0) -- cycle;
    \fill[pattern=north west lines, pattern color=haar4] (#2, 0) -- ++(#2,0) -- ++(0,#2) -- ++(-#2,0) -- cycle;
    \draw[draw=black, fill=none] (0, 0) -- ++(#2*2,0) -- ++(0,#2*2) -- ++(-#2*2,0) -- cycle;
    \node at(#2,0) [cmt,yshift=-0.3, anchor=north]{#3};
    \end{scope}}

  \def\layer#1#2#3#4{
    \begin{scope}[shift={(#1)}, line cap=round, thick]
    \draw[draw=black, fill=#3] (0, 0) -- ++(#2*2,0) -- ++(0,#2*2) -- ++(-#2*2,0) -- cycle;
    \node at(#2, 0) [cmt, yshift=-0.3, anchor=north]{{#4}};
  \end{scope}}

  \begin{tikzpicture}[inner sep=0pt]
  \layer{-1.4,-1.5}{0.5}{icra6!60}{conv5}
  \layer{6.0,-0.2}{0.5}{rainbow1!80}{}
  \node at(5.6, 0.3)[cmt,anchor=east]{conv\_pyr};
  \draw[arr](-0.4,-1.0)--++(0.4,0);
  \draw[arr,color=blue](-0.9,-0.5) --++(0,0.6)--++(6.9,0);

  \dwtlayer{0,-1.5}{0.5}{dwt\_p1}
  \layer{1.5,-1.4}{0.4}{haar1}{\hspace*{3em}$Y^{ll}_{p1}$}
  \layer{4.7,-1.4}{0.4}{rainbow1!40}{conv\_p1}
  \draw[arr](1.0,-1.0)--++(0.5,0);
  \draw[arr](2.3,-1.0)--++(2.4,0);
  \draw[arr] (5.5,-1.0)--++(0.8,0);
  \node at(5.8,-0.8) [cmt]{$\times2$};
  \draw[arr](1.9,-1.4)--++(0,-0.6);

  \dwtlayer{1.5,-2.8}{0.4}{}
  \layer{2.8,-2.6}{0.2}{haar1}{}
  \layer{4.9,-2.6}{0.2}{rainbow1!40}{conv\_p2}
  \draw[arr](2.3,-2.4)--++(0.5,0);
  \draw[arr](3.2,-2.4)--++(1.7,0);
  \draw[arr](3.0,-2.6)--++(0,-0.5);
  \draw[arr](5.3,-2.4)--++(1.2,0) --++(0,1.2);
  \node at(5.8,-2.2) [cmt]{$\times4$};

  \dwtlayer{2.8,-3.5}{0.2}{}
  \layer{3.6,-3.4}{0.1}{haar1}{}
  \layer{5.0,-3.4}{0.1}{rainbow1!40}{conv\_p3}
  \draw[arr](3.1,-3.3)--++(0.5,0);
  \draw[arr](3.8,-3.3)--++(1.2,0);
  \draw[arr](3.7,-3.4)--++(0,-0.6);
  \draw[arr,-](5.2,-3.3)--++(1.3,0) --++(0,1.5);
  \node at(5.8,-3.1) [cmt]{$\times8$};

  \dwtlayer{3.6,-4.2}{0.1}{}
  \layer{4.25,-4.15}{0.05}{haar1}{}
  \layer{5.05,-4.15}{0.05}{rainbow1!40}{conv\_p4}
  \draw[arr](3.8,-4.1)--++(0.45,0);
  \draw[arr](4.35,-4.1)--++(0.7,0);
  \draw[arr,-](5.15,-4.1)--++(1.35,0) --++(0,1.5);
  \node at(5.8,-3.9) [cmt]{$\times16$};

  \draw[thick, black] (6.5,-1.0) circle (0.2);
  \draw[thick, black] (6.3,-1.0) --++(0.4,0);
  \draw[thick, black] (6.5,-1.2) --++(0,0.4);
  \draw[arr] (6.5,-1.2) --++(0,1.0);
  \node at (6.9, -2.0)[cmt,anchor=west, rotate=90]{concatenate};

  \node at(2.8,-4.9)[text centered, font=\small,]{(a) wavelet pyramid variant: low frequency propagation (LFP)};
  \end{tikzpicture}

  \vspace{2ex}

  \begin{tikzpicture}
    \layer{-1.4,-1.5}{0.5}{icra6!60}{conv5}
    \dwtlayer{-0,-1.5}{0.5}{\hspace*{2em}dwt\_p1}
    \idwtlayer{6.1,-1.5}{0.5}{\hspace*{-4em}idwt\_p1}
    \highf{1.3,-1.4}{0.5}
    \node at(2.8, -0.7) [font=\small, anchor=west]{$Y^{lh}_{p1}, Y^{hl}_{p1}, Y^{hh}_{p1}$};
    \layer{6.1,0}{0.5}{rainbow1!80}{}
    \node at(5.8, 0.5)[cmt,anchor=east]{conv\_pyr};
    \draw[arr](-0.4,-1.0) --++(0.4,0);
    \draw[arr](0.5,-1.5) --++(0,-0.5);
    \draw[arr](1.0,-1.0) --++(5.1,0);
    \draw[arr](6.6,-0.5) --++(0,0.5);
    \draw[arr,color=blue](-0.9,-0.5) --++(0,0.7)--++(7.0,0);

    \layer{0.1,-2.8}{0.4}{haar1}{$Y^{ll}_{p1}$}
    \layer{1.3,-2.8}{0.4}{rainbow1!40}{conv\_p1}
    \dwtlayer{2.5,-2.8}{0.4}{}
    \idwtlayer{6.2,-2.8}{0.4}{}
    \highf{4.2,-2.7}{0.4}
    \draw[arr](0.9,-2.4) --++(0.4,0);
    \draw[arr](2.1,-2.4) --++(0.4,0);
    \draw[arr](2.9,-2.8) --++(0,-0.4);
    \draw[arr](3.3,-2.4) --++(2.9,0);
    \draw[arr](6.6,-2.0) --++(0,0.5);
    \draw[arr,color=blue](1.7,-2.0) --++(0,0.7)--++(4.4,0);

    \layer{2.7,-3.6}{0.2}{haar1}{}
    \layer{3.5,-3.6}{0.2}{rainbow1!40}{conv\_p2}
    \dwtlayer{4.3,-3.6}{0.2}{}
    \idwtlayer{6.4,-3.6}{0.2}{}
    \highf{5.2,-3.5}{0.2}
    \draw[arr](3.1,-3.4) --++(0.4,0);
    \draw[arr](3.9,-3.4) --++(0.4,0);
    \draw[arr](4.5,-3.6) --++(0,-0.4);
    \draw[arr](4.7,-3.4) --++(1.7,0);
    \draw[arr](6.6,-3.2) --++(0,0.4);
    \draw[arr,color=blue](3.7,-3.2) --++(0,0.6)--++(2.5,0);

    \layer{4.4,-4.2}{0.1}{haar1}{}
    \layer{5.0,-4.2}{0.1}{rainbow1!40}{conv\_p3}
    \dwtlayer{5.6,-4.2}{0.1}{}
    \highf{5.95,-4.1}{0.1}
    \idwtlayer{6.5,-4.2}{0.1}{}
    \draw[arr](4.6,-4.1) --++(0.4,0);
    \draw[arr](5.2,-4.1) --++(0.4,0);
    \draw[arr](5.8,-4.1) --++(0.7,0);
    \draw[arr](5.7,-4.2) --++(0,-0.4);
    \draw[arr](6.6,-4.0) --++(0,0.4);
    \draw[arr,color=blue](5.1,-4.0) --++(0,0.45)--++(1.3,0);

    \layer{5.65,-4.7}{0.05}{haar1}{}
    \layer{6.55,-4.7}{0.05}{rainbow1!40}{\hspace*{-1ex}conv\_p4}
    \draw[arr](5.76,-4.65) --++(0.75,0.0);
    \draw[arr](6.6,-4.6) --++(0.,0.4);

    \node at(2.8,-5.5)[text centered, font=\small,]{(b) wavelet pyramid variant: full frequency composition (FFC)};

  \end{tikzpicture}

  \caption{The proposed wavelet pyramid variants, with the data flow indicated by black arrows and shortcuts by blue arrows. Both pyramids take conv5 as input and produce conv\_pyr as output, without changing the spatial resolution. Both pyramids build a multi-resolution wavelet pyramid via successive DWT. The LFP pyramid only utilizes the low-low frequency $Y^{ll}$, where each $Y^{ll}$ is filtered by further convolutions, bilinear upsampled to the input resolution and concatenated. The FFC pyramid employs the high frequency bands for upscaling via iDWT.}

  \label{arxiv18:fig:pyramids}

\end{figure}

\subsubsection{Wavelet Unpooling}\label{sect:waveletunpool}
WCNN achieves the unpooling through iDWT layers. To this end, the DWT layers are added consistently into the encoder to obtain high-frequency components. The idea is straight-forward. At encoder, the DWT layers decompose the feature map into four frequency bands channelwise, where each frequency band is half-resolution of the input. The high-frequency components are skip-connected to the decoder where the spatial resolution needs to be upscaled by a factor of two. Taking the layer idwt\_4 in \cref{arxiv18:fig:network} as an example, the input to this layer are four components of spatial resolution $1/32$ to perform iDWT. The pyramid output serve the low-low frequency $\tY^{ll}$, while the output of the dwt4 layer operating on the conv4 provide the three high-frequency components $Y^{lh}$, $Y^{hl}$, and $Y^{hh}$. With iDWT, the spatial resolution is upscaled to $1/16$. The output of layer idwt4 is finalized by adding the $1/16$ resolution direct output of conv4, which is a standard skip connection commonly used by many state-of-the-art encoder-decoder CNNs to improve the upsampling performance. The iDWT layer can thus be described by
\begin{equation}\label{eq:idwtlayer}
  G_h^{-1}(\tY^{ll}, Y^{lh}, Y^{hl}, Y^{hh}) + X \;.
\end{equation}
We denote this appproach of upscaling the decoder feature map with the wavelet coefficients from the encoder as wavelet unpooling.

Typically, CNNs extract feature with many layers of convolution and nonlinear operations, which transform and embed the feature space differently layer by layer. The wavelet unpooling aims to maintain the similar frequency structure throughout CNNs. By replacing the low-frequency of the encoder with the corresponding output of the decoder to perform iDWT with the high-frequency bands from the encoder, the wavelet unpooling aims to enforce learning feature maps of  invariant frequency structure under layers of filtering. The skip connections of the signals before DWT also support learning such consistency.

In comparison to the other unpooling methods, for example to upsampling by transposed convolution as proposed in \cite{cnn:long15cvpr:fcn}, wavelet unpooling does not require any parameters for both DWT and iDWT layers. Compare to the memorized unpooling as proposed in \cite{cnn:zeiler11cvpr:memorizedunpool}, or the method to map the low-resolution feature map to the top-left entry of a $2\times 2$ block~\cite{cnn:dosovitskiy15cvpr:chairs}, the wavelet unpooling aims to restore every entries according to the frequency structure.


\subsubsection{Wavelet Pyramid}\label{sect:pyramid}
With CNNs that are designed for classification task, the last few layers typically reduce the spatial resolution to $1\times1$. Such feature maps have the receptive field of the entire input image and therefore capture the global context. Recent works have demonstrated that capturing global context information is also crucial for accurate dense pixelwise prediction~\cite{cnn:chen15iclr:deeplab, cnn:zhao17cvpr:pspnet}. While it is straight-forward to obtain global context with fully-connected layer or with convolutional layers of large filter size, it is difficult to bridge an encoder with drastically reduced spatial resolution to a proper decoder. Most state-of-the-art CNN encoder reduce the spatial resolution by a factor of 32, which produces $7\times7$ output given $224\times224$ input dimensions. If the global context is captured by a simple fully-connected layer, learning $7\times7$
upsampling kernels is challenging.

One solution is to use the dilated convolutions, which increase the perceptive field with the same amount of parameters~\cite{cnn:chen15iclr:deeplab, cnn:chen18:deeplabv2}. Building on the dilated CNNs, the pyramid spatial pooling network PSPNet~\cite{cnn:zhao17cvpr:pspnet} introduces a pyramid on the feature map with multiple average pooling of different window sizes. Noticeably, the dilated convolutions demand considerably larger amounts of memory to host the data, which quickly becomes the bottleneck for training with large batch size. In this work, we base our network design on non-dilated CNNs and instead construct the pyramids through wavelet transformations. We propose two wavelet pyramids variants, namely the low frequency propagation (LFP) and the full frequency composition (FFC) as shown in \cref{arxiv18:fig:pyramids}.

\paragraph{Low-Frequency Propagation Wavelet Pyramid}
Shown in \cref{arxiv18:fig:pyramids}~(a), the LFP pyramid successively performs DWT on the low-low frequency components $Y^{ll}$. At each pyramid level, the extracted $Y^{ll}$ component is further transformed with a convolutional layer, which is then bilinear upsampled to the same spatial resolution as the pyramid input, i.e., conv5. We then concatenate these the upsampled feature maps to aggregate the global context that are captured at different scale. This concatenated feature map is combined with the skip-connected conv5 by an elementwise addition, which sis then filtered with a $1\times1$ convolutional layer to match the channel dimension of the decoder.

With LFP, a multi-resolution wavelet pyramid is constructed, where only the low-low frequency bands of each level are used. The LFP pyramid resembles the pyramid proposed by the PSPNet~\cite{cnn:zhao17cvpr:pspnet}. In particular, with the Haar wavelet, the low-low frequency is equivalent to the average pooling by a $2\times2$ window. However, the difference is the PSPNet design average pooling with a multiple heuristic window size, whereas LFP pyramid is strictly performed accordingly to frequency decomposition. Despite the Haar wavelet is used in this work, the LFP pyramid can be easily generalized with other wavelet base functions.

\paragraph{Full-Frequency Composition Wavelet Pyramid}
The LFP pyramid only uses the low-low frequency bands. In order to make full use of the frequency decomposition, the FFC pyramid is developed. Shown in \cref{arxiv18:fig:pyramids}~(b), the FFC pyramid amounts to a small encoder-decoder with wavelet unpooling. Start from the input conv5, DWT is performed to obtain the four frequency bands. The low-low frequency band $Y^{ll}$ is filtered by an additional convolutional layer and the high frequency bands $Y^{lh}, Y^{hl}, Y^{hh}$ are cached for unpooling. The filtered low-low frequency is then further decomposed by DWT into the finer level and the same operation repeats until the finest feature map is obtained. To upscale from the finest level, we again adopt the wavelet unpooling as described by \cref{eq:idwtlayer}. To this end, the iDWT is first performed using the cached high frequency bands, and then the output is further fused with the skip connection. The wavelet unpooling successively restore the spatial resolution to the same as the input to the pyramid. Finally, we skip connect conv5 with the wavelet output by an elementwise addition, and project the global context with a $1\times1$ convolution to bridge the following decoder. It can be seen that, the FFC pyramid mimic the encoder-decoder design, which naturally reduces the spatial resolution and restore it in the consistent manner with the remaining network.

\section{Evaluations}
In this section, we evaluate the proposed WCNN method for the task of semantic image segmentation. To this end, we use the Cityscape benchmark dataset~\cite{dataset:cordts16cvpr:cityscapes} which contains 2,975 training, 500 validation and 1,525 test images that are captured in 50 different cities from a driving car. All the images are densely annotated into 30 commonly observed objects classes occurring in urban street scenes from which 19 classes are used for evaluation. The Cityscape benchmark provides all the images with the same high resolution of~$2048\times 1024$. The ground truth for the test images is not publicly available and evaluations on the test set are submitted online\footnote{http://www.cityscapes-dataset.com} for fair comparison.

\begin{table}
  \centering
  \caption{The layer configurations of the proposed WCNN (see \cref{arxiv18:fig:network}). The encoder is based on ResNet101~\cite{cnn:he16cvpr:resnet}. The resblock is the residual block from ResNet, where $(x, y)\times z$ denotes stacking $z$ blocks of $[(1\times1, x), (3\times3, x), (1\times1, y)]$ convolutional layers. For upconvolution, the transposed convolution is first used to upscale the input by a factor of two, followed by residual blocks. We denote the stride-two operations with s2, and elementwise addition with $\boxplus$. The dimension of the layer output assumes the spatial resolution of input image is normalized to 1, and the second entry denotes the depth of the feature maps.}
  \label{arxiv18:tab:wcnn}
  \setlength{\tabcolsep}{3pt}
  \begin{tabular}{L{10ex}L{25ex}L{17ex}L{10ex}}
    \toprule
    layer    & operation & input & dimension \\
    \midrule
    conv1    & $(7\times7, 64)$, s2           &RGB        &1/2, 64\\
    maxpool  & $(2\times2)$, s2               &conv1      &1/4, 64\\
    conv2\_x & resblock $(64, 256)\times3$    &maxpool    &1/4, 256 \\
    dwt2     & $G_h$                          &conv2\_x   &1/8, 256\\
    conv3\_1 & resblock $(128, 512)$, s2      &conv2\_x   &1/8, 512\\
    conv3\_x & resblock $(128, 512)\times3$   &conv3\_1   &1/8, 512\\
    dwt3     & $G_h$                          &conv3\_x   &1/16, 512\\
    conv4\_1 & resblock $(256,1024)$, s2      &conv3\_x   &1/16, 1024\\
    conv4\_x & resblock $(256,1024)\times22$  &conv4\_1   &1/16, 1024\\
    dwt4     & $G_h$                          &conv4\_x   &1/32, 1024\\
    conv5\_1 & resblock $(512,2048)$, s2      &conv4\_x   &1/32, 2048\\
    conv5\_x & resblock $(512,2048)\times2$   &conv5\_1   &1/32, 2048\\
    \midrule
    pyramid  & &conv5x   &1/32, 1024\\ 
    \midrule
    idwt4     & $G_h^{-1}$                 &$\begin{dcases}\text{pyramid} \\ Y_4^{lh}, Y_4^{hl}, Y_4^{hh}\end{dcases}$ &1/16, 1024\\
    dconv4\_x & resblock $(256,512)\times3$   &idwt4 $\boxplus$ conv4\_x &1/16, 512\\
    idwt3     & $G_h^{-1}$                 &$\begin{dcases}\text{dconv4\_x}\\ Y_3^{lh}, Y_3^{hl}, Y_3^{hh}\end{dcases}$ &1/8, 512\\
    dconv3\_x & resblock $(128,256)\times3$    &idwt3 $\boxplus$ conv3\_x &1/8, 256\\
    idwt2     & $G_h^{-1}$                 &$\begin{dcases}\text{dconv3\_x}\\ Y_2^{lh}, Y_2^{hl}, Y_2^{hh}\end{dcases}$ &1/4, 256\\
    dconv2\_x & resblock $(64,128)\times3$    &idwt2 $\boxplus$ conv2\_x &1/4, 128\\
    upconv2\_x& upconv $(64, 64)\times3$      &dconv2\_x &1/2, 64\\
    upconv1\_x& upconv $(64, 64)\times2$      &upconv2\_x &1/1, 64\\
    \bottomrule
  \end{tabular}
\end{table}

\subsection{WCNN Configurations}
\Cref{arxiv18:tab:wcnn} presents the network configurations of the proposed WCNN. We take the state-of-the-art ResNet101~\cite{cnn:he16cvpr:resnet} for the encoder. The ResNet101 uses stride-two convolution to reduce spatial resolution. To implement WCNN, we preserve the stride-two convolution layers and insert three DWT layers dwt2, dwt3, dwt4 into the decoder conv2\_x, conv3\_x, conv4\_x, respectively to obtain the frequency bands. At each upscaling stage at the decoder, the corresponding frequency bands are used, then followed by several residual blocks before the next upscaling stage. The last two upscaling stages are implemented as upconvolution, where transposed convolution is first applied to scale up the resolution by a factor of two, then residual blocks are used to further filter the intermediate output. In WCNN, we reply heavily on the residual blocks proposed in ResNet \cite{cnn:he16cvpr:resnet}, where each block is a stack of three convolutional layers with the second layer of $3\times3$ for feature extraction and the first and third layers as $1\times1$ convolutions for feature projection.

In this work, we develop CNNs for high-resolution predictions. An input image of $512\times1024$ yields conv5\_x to have the spatial resolution of $16\times32$. Therefore, we design both LFP and FFC pyamids to have four levels of DWT, which produce the four levels of frequency components of $8\times16$, $4\times8$, $2\times4$ and $1\times2$, respectively. The finest pyramid level thus has the receptive field of the entire input. The details of the LFP and FFC pyramids are given in \cref{arxiv18:tab:pyr}.

To evaluate the proposed network, the baseline CNN is designed to have minimum difference with WCNN. Taking the WCNN configuration in \cref{arxiv18:tab:wcnn}, the baseline model 1) removes all DWT layers at encoder 2) replaces the pyramid by one $1\times1, 1024$ convolutional layer, and 3) replaces the iDWT layers by transposed convolution to upscale the feature map by a factor of 2. The rest layers are the same with WCNN. In the following experiment, we compare the baseline model, the baseline model with LFP and FFC pyramid, the WCNN with LFP and FFC pyramids.

\begin{table}
  \centering
  \caption{The configurations of the proposed LFP and FFC pyramids (see \cref{arxiv18:fig:pyramids}). Assuming conv5 has a resolution of $16\times32, 2048$, both LFP and FFC pyramids have four levels. For simplicity, the outer two levels are presented in the table, whereas the inner two levels repeats the same patterns. The operator $\star a$ denotes bilinear upsample by a factor of $a$ and the operator $\boxplus$ denotes elementwise addition. }
  \label{arxiv18:tab:pyr}
  \setlength{\tabcolsep}{3pt}
  \begin{tabular}{L{10ex}L{17ex}L{20ex}L{13ex}}
    \toprule
    \multicolumn{4}{c}{LFP-pyramid} \\
    layer    & operation & input & dimension \\
    \hline
    dwt\_p1  & $G_h$                          &conv5          &$8\times16$, 2048\\
    conv\_p1 & $(1\times1, 512)$              &$Y^{ll}_{p1}$  &$8\times16$, 512\\
    dwt\_p2  & $G_h$                          &$Y^{ll}_{p1}$  &$4\times8$,  512\\
    conv\_p2 & $(1\times1, 512)$              &$Y^{ll}_{p2}$  &$4\times8$, 512\\
    $\vdots$ &$\vdots$ &$\vdots$&$\vdots$\\
    concat   & concatenation                  &$\begin{dcases}Y^{ll}_{p1}\star2, Y^{ll}_{p2}\star4 \\Y^{ll}_{p3}\star8, Y^{ll}_{p4}\star16 \end{dcases}$  &$16\times32$,2048\\
    conv\_pyr &$(1\times1, 1024)$      &concat $\boxplus$ conv5 &$16\times32$,1024\\
    \midrule
    \multicolumn{4}{c}{FFP-pyramid} \\
    layer    & operation & input & dimension \\
    \hline
    dwt\_p1  & $G_h$                          &conv5          &$8\times16$, 2048\\
    conv\_p1 & $(1\times1, 2048)$             &$Y^{ll}_{p1}$  &$8\times16$, 2048\\
    dwt\_p2  & $G_h$                          &conv\_{p1}     &$4\times8$,  2048\\
    conv\_p2 & $(1\times1, 2048)$             &$Y^{ll}_{p2}$  &$4\times8$, 2048\\
    $\vdots$ &$\vdots$ &$\vdots$&$\vdots$\\
    idwt\_p2     & $G_h^{-1}$                      &$\begin{dcases}\text{conv\_p2}\boxplus\text{idwt\_p3}\\ Y_2^{lh}, Y_2^{hl}, Y_2^{hh}\end{dcases}$ &$8\times16$, 2048\\
    idwt\_p1     & $G_h^{-1}$                      &$\begin{dcases}\text{conv\_p1}\boxplus\text{idwt\_p2}\\ Y_1^{lh}, Y_1^{hl}, Y_1^{hh}\end{dcases}$ &$16\times32$, 2048\\
    conv\_pyr &$(1\times1, 1024)$      &idwt\_p1$\boxplus$ conv5 &$16\times32$, 1024\\
    \bottomrule
  \end{tabular}
\end{table}

\subsection{Implementation Details}
We have implemented all our methods based on the TensorFlow~\cite{cnn:martin16:tensorflow} machine learning framework. For network training, we initialize the parameters of the encoder layers from pretrained ResNet model on ImageNet and initialize the convolutional kernels on the decoder with He~\cite{cnn:he15iccv:msrcinit} initialization. We run the training with batch size of four on the Nvidia Titan X GPU. For both training, we minimize the cross-entropy loss using the Stochastic Gradient Descent (SGD) solver with Momentum of 0.9. The initial learning rate is set to 0.001 and decrease with a factor of 0.9 every 10 epoch. We train the network until convergences. For cityscapes, all the variants of our experiments converges around 60K iterations. Following~\cite{cnn:pohlen17cvpr:frrn}, we apply bootstrapping loss minimization for Cityscapes benchmark in order to speed up the training and boost the segmentation accuracy. For all Cityscapes experiments, we fix the threshold of bootstrapping to the top 8192 most difficult pixels per images.

\begin{table*}
    \centering
    \caption{Cityscapes 19-class semantic segmentation IoU scores on \emph{val} set. All test results are obtained by comparing to half resolution ground-truth labeling, which is the resolution of input images into our networks. The second part of the table report the performance with multi-scale test time data augmentation, indicated by the MS suffix.}
    \label{tab:classwiseiou}
    \setlength{\tabcolsep}{2.5pt}
    \begin{tabular}{L{20ex}  cl*{20}{r} }
        \toprule
        method
        &\block{city0}{road}
        &\block{city1}{sidewalk}
        &\block{city2}{building}
        &\block{city3}{wall}
        &\block{city4}{fence}
        &\block{city5}{pole}
        &\block{city6}{traffic}
        &\block{city7}{traffic light}
        &\block{city8}{vegetarian}
        &\block{city9}{terrain}
        &\block{city10}{sky}
        &\block{city11}{person}
        &\block{city12}{rider}
        &\block{city13}{car}
        &\block{city14}{truck}
        &\block{city15}{bus}
        &\block{city16}{train}
        &\block{city17}{motorcycle}
        &\block{city18}{bicycle}
        &avg\\
        \midrule
        frequency &37.7 & 5.4 &21.9 & 0.7 & 0.8 & 1.5 & 0.2 & 0.7 &17.2 & 0.8 & 3.4 & 1.3 & 0.2 & 6.6 & 0.3 & 0.4 & 0.1 & 0.1 & 0.7 \\
        \midrule
        baseline & 98.8 &88.8 &96.0 &51.5 &61.6 &62.0 &66.6 &76.5 &96.0 &70.1 &97.1 &85.8 &66.4 &97.0 &81.4 &85.4 &59.0 &53.8 &84.6 & 69.2\\
        baseline-LFP &98.6 &90.1 &95.5 &62.6 &62.6 &61.3 &65.7 &76.0 &95.9 &69.3 &97.4 &85.4 &63.6 &97.1 &80.1 &88.4 &73.8 &61.2 &85.1 &71.2\\
        baseline-FFC &98.6 &89.6 &95.3 &63.4 &62.0 &61.3 &67.8 &74.4 &96.1 &64.6 &97.3 &85.9 &63.0 &96.9 &85.5 &89.4 &73.6 &58.5 &84.5 &70.7\\
        WCNN-LFP &98.6 &89.8 &95.7 &63.0 &65.8 &61.5 &67.8 &76.2 &96.3 &69.4 &97.4 &85.8 &67.4 &97.2 &82.0 &88.9 &69.9 &59.9 &84.9 &71.6\\ 
        WCNN-FFC &98.7 &90.5 &95.6 &64.8 &64.6 &63.2 &67.8 &77.3 &96.1 &71.0 &97.3 &86.1 &65.3 &97.0 &82.7 &88.7 &77.6 &57.7 &85.1 &71.9\\ 
        \midrule
        baseline-MS &\bf 99.0 &90.6 &\bf 96.7 &48.0 &61.2 &68.2 &72.9 &80.2 &96.3 &\bf72.5 &97.7 &89.1 &70.3 &97.6 &76.6 &82.2 &48.9 &60.7 &84.9 &71.4\\
        baseline-LFP-MS &98.7 &92.2 &96.5 &54.0 &65.5 &68.9 &71.2 &79.0 &96.1 &64.7 &97.6 &88.1 &64.3 &\bf 97.8 &71.2 &87.3 &71.8 &\bf 68.5 &85.7 &73.3\\
        baseline-FFC-MS &98.7 &91.7 &96.4 &64.6 &65.0 &67.4 &\bf 74.3 &79.7 &\bf 96.7 &68.9 &\bf 98.0 &88.8 &68.9 &97.5 &\bf 88.3 &\bf 90.6 &\bf 79.3 &60.9 &\bf 85.8 &74.7\\
        WCNN-LFP-MS &98.8 &\bf 92.4 &96.2 &61.2 &\bf 68.0 &68.5 &71.2 &79.8 &96.3 &64.8 &97.5 &88.4 &\bf 70.1 &\bf 97.8 &77.8 &89.3 &61.6 &74.1 &87.1 &73.9\\
        WCNN-FFC-MS &98.8 &92.2 &96.6 &\bf 68.6 &64.8 &\bf 69.1 &73.9 &\bf 81.6 &\bf 96.7 &72.4 &97.8 &\bf 89.3 &68.9 &97.5 &87.3 &90.5 &73.3 &58.0 &85.3 &\bf 75.2\\
        \bottomrule
    \end{tabular}
\end{table*}

\begin{table}
  \centering
  \caption{IoU scores for the Cityscapes 19-class and category semantic segmentation on the \emph{test} set (benchmark). All test results are obtained by testing on half resolution and comparing to full resolution groundtruth labeling through upsampling.}
  \label{tab:testsetmiou}
  \setlength{\tabcolsep}{1em}
  \begin{tabular}{L{20ex}L{18ex}L{18ex}}
  \toprule
  method  &class mIoU &category mIoU\\
  \midrule
  FRRN~\cite{cnn:pohlen17cvpr:frrn} & 71.8 & \bf 88.9\\
  WCNN-FFC & 70.9 & 86.1 \\
  WCNN-FFC-MS & \bf 73.7 & 88.3 \\
  \bottomrule
  \end{tabular}
\end{table}

\begin{figure*}
\centering
\begin{tikzpicture}[inner sep=1pt]
    \def\imwidth{12em}
    \def\imh{6em}
    \def\imsep{1mm}
    \tikzstyle{cmt}=[rotate=90,anchor=center, font=\normalsize, text centered]

    \def\onecmp#1#2#3
    {
    \begin{scope}
        \node(rgb) at(#2,0)[anchor=west]{\includegraphics[width=\imwidth]{#1_leftImg8bit.png}};
        \node(gt)  at(rgb.south)[anchor=north]{\includegraphics[width=\imwidth]{#1_gtFine.png}};
        \node(uppsp)  at(gt.south)[anchor=north,,yshift=-3pt] {\includegraphics[width=\imwidth]{#1_pred_upconv_psp.png}};
        \node(uppsp2) at(uppsp.south)[anchor=north] {\includegraphics[width=\imwidth]{#1_pred_residual_upconv_psp.png}};
        \node(uphaar)  at(uppsp2.south)[anchor=north,yshift=-3pt] {\includegraphics[width=\imwidth]{#1_pred_upconv_haar.png}};
        \node(uphaar2) at(uphaar.south)[anchor=north] {\includegraphics[width=\imwidth]{#1_pred_residual_upconv_haar.png}};
        \node(wcnnpsp)  at(uphaar2.south)[anchor=north,yshift=-3pt] {\includegraphics[width=\imwidth]{#1_pred_wcnn_psp.png}};
        \node(wcnnpsp2) at(wcnnpsp.south)[anchor=north] {\includegraphics[width=\imwidth]{#1_pred_residual_wcnn_psp.png}};
        \node(wcnnhaar)  at(wcnnpsp2.south)[anchor=north,yshift=-3pt] {\includegraphics[width=\imwidth]{#1_pred_wcnn_haar.png}};
        \node(wcnnhaar2) at(wcnnhaar.south)[anchor=north] {\includegraphics[width=\imwidth]{#1_pred_residual_wcnn_haar.png}};
    \end{scope}
    }
    \onecmp{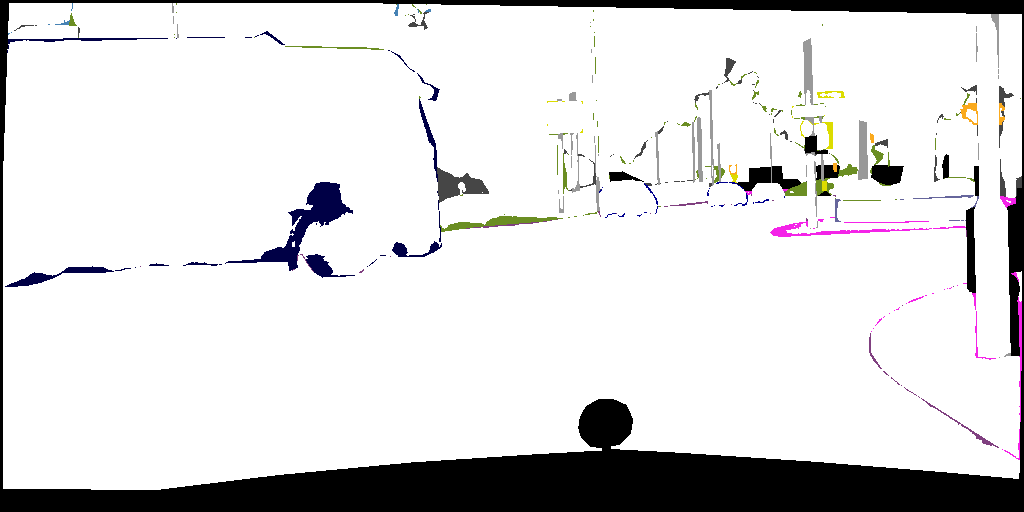}{0}
    \onecmp{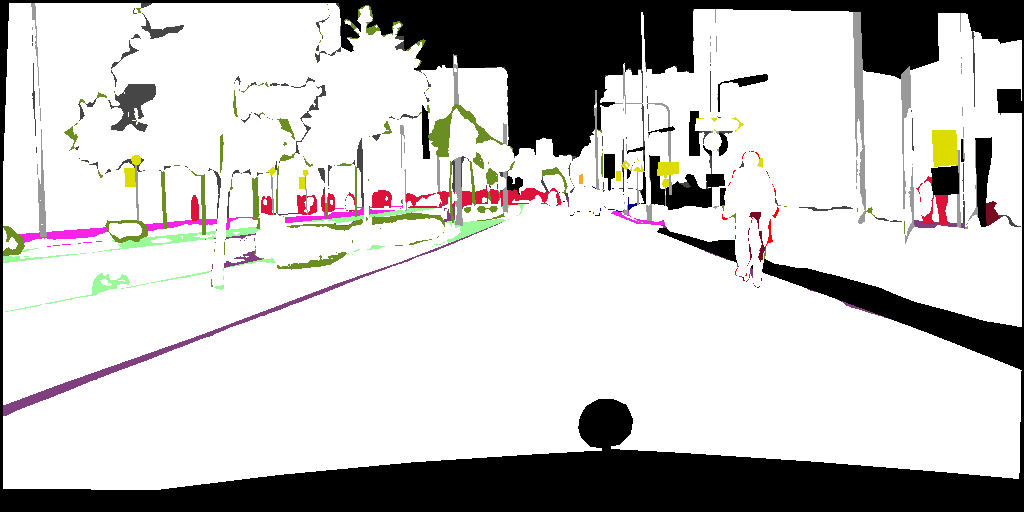}{\imwidth+\imsep}
    \onecmp{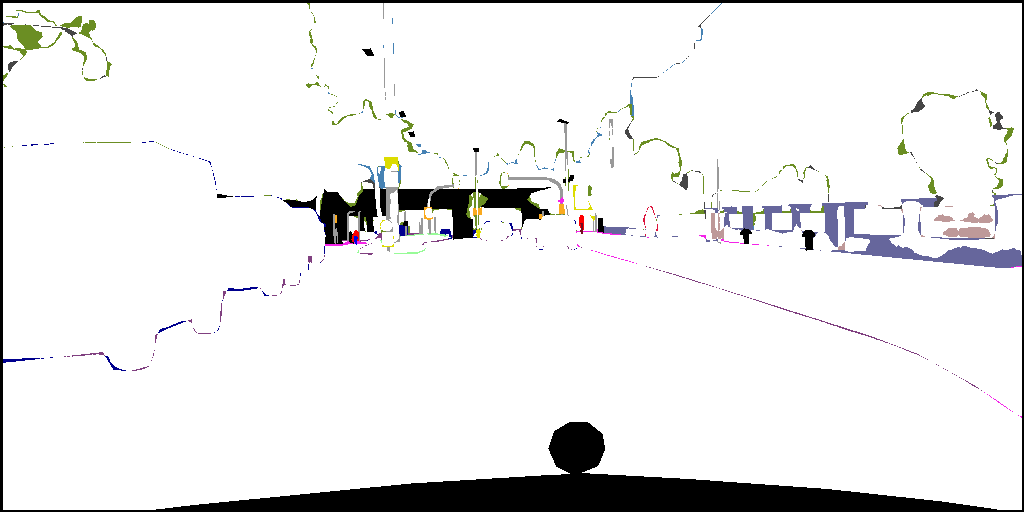}{\imwidth*2+\imsep*2}
    \onecmp{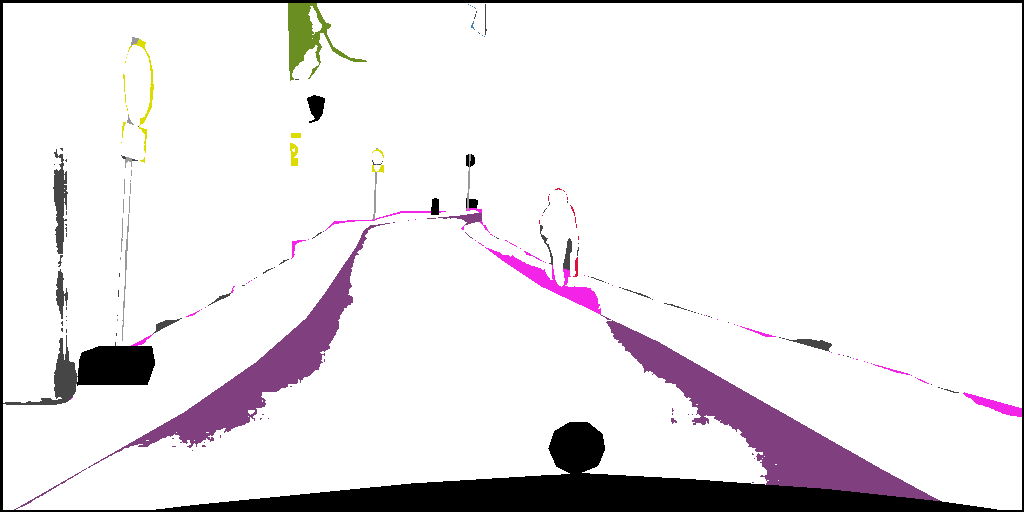}{\imwidth*3+\imsep*3}

    \tikzstyle{cmt}=[font=\normalsize, text centered, anchor=east, rotate=90]
    \begin{scope}[xshift=-2ex]
      \node(t1) at (0,0)[text width=\imh, yshift=2.5em, xshift=0em, cmt]{RGB};
      \node(t2) at (t1.west)[text width=6em, yshift=0.5em, xshift=0em, cmt]{ground truth};
      \node(t3) at (t2.west)[text width=2*\imh, yshift=-0em, xshift=0em, cmt]{baseline-LFP-MS};
      \node(t4) at (t3.west)[text width=2*\imh, yshift=-0em, xshift=0em, cmt]{baseline-FFC-MS};
      \node(t5) at (t4.west)[text width=2*\imh, yshift=-1em, xshift=0em, cmt]{WCNN-LFP-MS};
      \node(t6) at (t5.west)[text width=2*\imh, yshift=-0.5em, xshift=0em, cmt]{WCNN-FFC-MS};
    \end{scope}
\end{tikzpicture}
\caption{Qualitative exemplary semantic segmentation results on the Cityspaces dataset. From top to bottom: RGB image, ground-truth segmentation, baseline-LFP-MS, baseline-FFC-MS, WCNN-LFP-MS, WCNN-FFC-MS. The semantic color coding is given in \cref{tab:classwiseiou}.}
\label{fig:cityscapes}
\end{figure*}
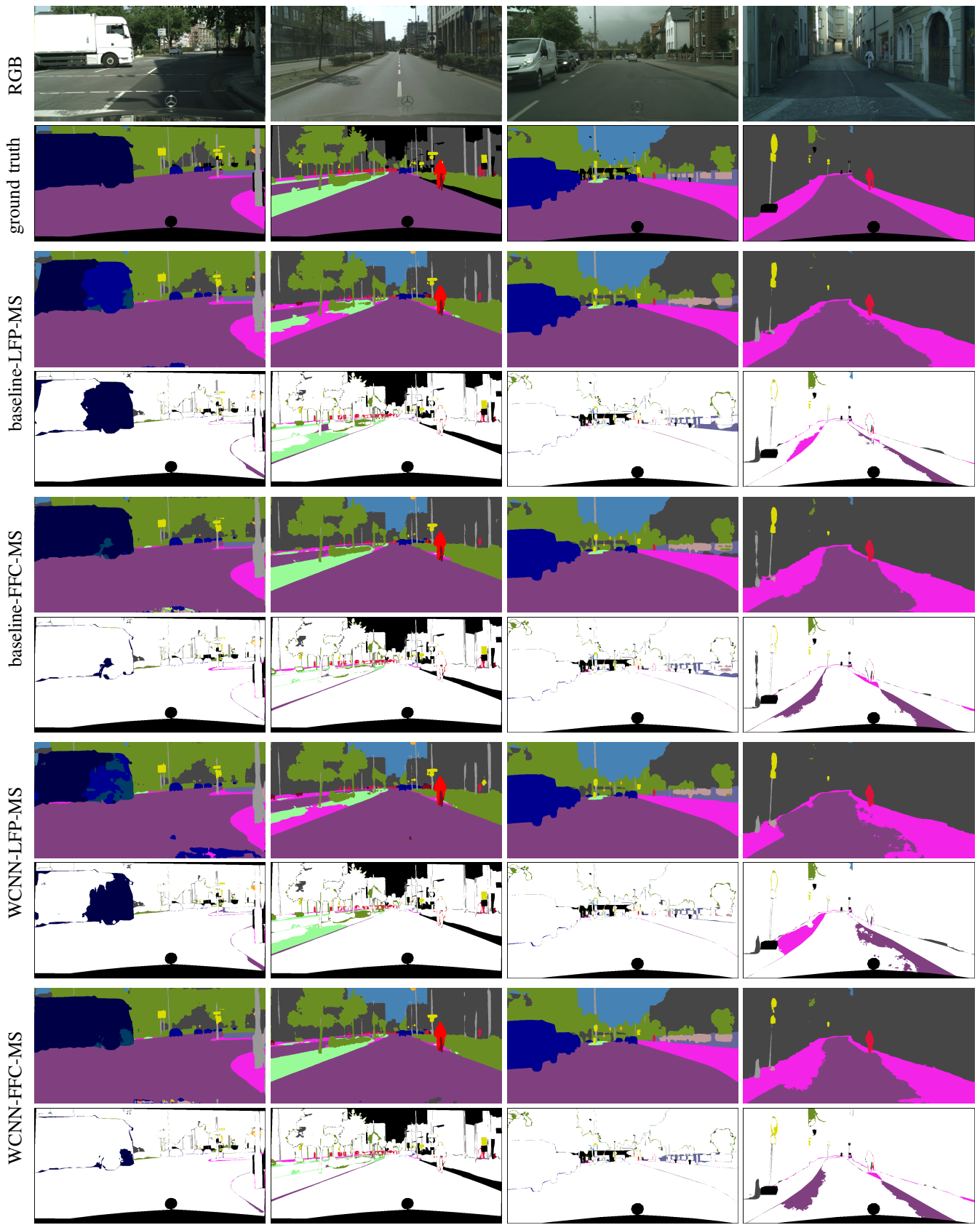

To train all the variants of the baseline and our model, we fix the input to the network to quarter resolution of the original dataset, i.e., $512\times 1024$. For evaluation on the validation dataset, we upsample the output logits bilinear to half of the resolution (to match the network input resolution) and compute the intersection-over-union (IoU) score for each class and on average. We also experiment with test time data augmentation, where we randomly scale the input images and feed them through the network before fuse the score.

\subsection{Cityscapes}
We evaluate segmentation accuracy using the commonly used evaluation metric of IoU. \Cref{tab:classwiseiou} gives the class-wise IoU and the mean IoU over the 19 classes. It can be seen that adding LFP and FFC pyramids to the baseline network already significantly improves the segmentation performance over the baseline. The FFC pyramid consistently outperforms the LFP pyramid. With WCNN we gain another increase in mean IoU of up to 1.2 over the corresponding baseline. With multi-scale test time augmentation, the accuracy of each model is increased, but the similar rank is observed among the different methods. Our variants strongly benefit, while the combination of wavelet unpooling and FFC wavelet pyramid achieves best increase in performance towards the baseline (6.0 mIoU). These results demonstrate that wavelet unpooling as well as the FFC wavelet pyramid improve the dense prediction of the baseline model. The qualitative comparisons are shown in \cref{fig:cityscapes}. It can be seen that the WCNN approach recovers fine-detailed structures such as fences, poles or traffic signs with higher accuracy than the baselines.

\Cref{tab:testsetmiou} compares our method with the current state-of-the-art method FRRN~\cite{cnn:pohlen17cvpr:frrn} on the same input resolution (2x subsampling) on the Cityscapes benchmark. It can be seen that our method WCNN-FFC-MS outperforms FRRN by 1.9 mean IoU over the 19-classes while it is worse (0.6 mIoU) on the category level. Notably, WCNN is much less memory demanding than FRRN.

\section{Conclusion}

This paper introduce WCNN, a novel encoder-decoder CNN architecture for dense pixelwise prediction. The key innovation is to exploits the discrete wavelet transform (DWT) and inverse DWT to design the unpooling operation. In the proposed network, the high-frequency coefficients extracted by DWT at the encoder stage are cached and later combined with coarse-resolution feature maps at the decoder to perform accurate upsampling and hence, ultimate pixelwise prediction. Further, two wavelet pyramid variants are introduced, i.e., the low frequency propagation (LFP) pyramid and the full frequency composition (FFC) pyramid. Both pyramid extract the global context from the encoder output with multi-resolution wavelet decomposition. Shown in experiment, WCNN outperforms the variant baseline CNNs and achieve the state-of-the-art semantic segmentation performance on the Cityscapes dataset.

In the future work, we will evaluate WCNNs for different dense pixelwise prediction tasks, e.g., depth estimation and optical flow estimation. We will also perform ablation study of the wavelet pyramid to evaluate different pyramid configuration. It is also interesting to extend the WCNN for different wavelet base functions or ultimately learn the optimal base functions with CNNs.

\balance
\bibliographystyle{ieeetr}
\bibliography{ms}

\end{document}